\documentclass[letterpaper, 10 pt, conference]{ieeeconf}  

\IEEEoverridecommandlockouts                              

\usepackage{graphicx} 
\usepackage{float} 
\usepackage{subfigure} 
\usepackage{amsmath}
\usepackage{mathtools} 
\usepackage{booktabs} 
\usepackage{array}    
\usepackage{multirow} 
\usepackage{caption}
\usepackage{cite}

\title{\LARGE \bf
V2I-Calib: A Novel Calibration Approach for Collaborative Vehicle and Infrastructure LiDAR Systems
}

\author{Qianxin Qu$^{\dag}$$^{1,2}$, Yijin Xiong$^{\dag}$$^{1,2}$, Guipeng Zhang$^{3}$, Xin Wu$^{2}$ \\ Xiaohan Gao$^{2}$,  Xin Gao$^{2}$, Hanyu Li$^{2}$, Shichun Guo$^{2}$, and Guoying Zhang$^{*}$$^{1}$\\
$^{\dag}$Equal Contribution \qquad 
$^{*}$ Corresponding Author 
\thanks{Qianxin Qu$^{\dag}$ and Yijin Xiong$^{\dag}$ contributed to the work equllly and should be regarded as co-first authors.}
\thanks{This work was supported by the National Natural Science Foundation of China under Grant No. 52204180 and Central University Basic Research Fund of China under Grant No. 2024CXCYZ010.}
\thanks{$^{1}$the School of Artificial Intelligence, China University of Mining and Technology -Beijing, Beijing, 100083, China. }
\thanks{$^{2}$the State Key Laboratory of Automotive Safety and Energy, and  School of Vehicle and Mobility, Tsinghua University, Beijing, 100084,  China. }
\thanks{$^{3}$Computer Science and Technology in Institute of Computing Technology, Chinese Academy of Sciences, Beijing, China. }
}

\begin{document}

\maketitle
\thispagestyle{empty}
\pagestyle{empty}


\begin{abstract}

Cooperative LiDAR systems integrating vehicles and road infrastructure, termed V2I calibration, exhibit substantial potential, yet their deployment encounters numerous challenges. A pivotal aspect of ensuring data accuracy and consistency across such systems involves the calibration of LiDAR units across heterogeneous vehicular and infrastructural endpoints. This necessitates the development of calibration methods that are both real-time and robust, particularly those that can ensure robust performance in urban canyon scenarios without relying on initial positioning values. Accordingly, this paper introduces a novel approach to V2I calibration, leveraging spatial association information among perceived objects. Central to this method is the innovative Overall Intersection over Union (oIoU) metric, which quantifies the correlation between targets identified by vehicle and infrastructure systems, thereby facilitating the real-time monitoring of calibration results. Our approach involves identifying common targets within the perception results of vehicle and infrastructure LiDAR systems through the construction of an affinity matrix. These common targets then form the basis for the calculation and optimization of extrinsic parameters. Comparative and ablation studies conducted using the DAIR-V2X dataset substantiate the superiority of our approach. For further insights and resources, our project repository is accessible at https://github.com/MassimoQu/v2i-calib.

\end{abstract}


\section{INTRODUCTION}


Perception systems are crucial for ensuring the safe and effective operation of autonomous vehicles \cite{autonomousdrivingIROS2023}. Vehicle-to-infrastructure (V2I) cooperative systems can enhance the reliability of perception systems in complex traffic scenarios or adverse weather conditions by complementing vehicle-end and road-end information \cite{badweathersensors2024}. Collaborative vehicle and roadside LiDAR systems hold significant potential, yet their realization faces numerous challenges, especially in the calibration of sensor systems.

Calibration of LiDAR systems across heterogeneous vehicle and infrastructure endpoints is a vital step to ensure the accuracy and consistency of perception system data. This requires high-precision synchronization and registration techniques to integrate LiDAR data from both vehicles and infrastructure \cite{cooperativeIROS2023}. This not only involves complex technical issues such as time synchronization, spatial alignment, and data fusion but also necessitates real-time processing capabilities in dynamic environments.

\begin{figure}[]
	\centering
	\vspace{3mm}
	\setcounter{subfigure}{0}
 \subfigure[Vehicle Perspective]{
    \begin{minipage}[t]{0.52\linewidth}
        \centering
        \includegraphics[width=.8\linewidth]{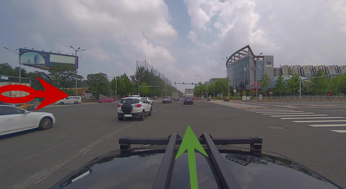}  
        \label{fig:sub1}
    \end{minipage}
}       
\hspace{-12mm}
\subfigure[Infrastructure Perspective]{
    \begin{minipage}[t]{0.52\linewidth}
        \centering
        \includegraphics[width=.8\linewidth]{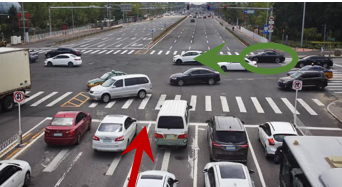}  
        \label{fig:sub2}
    \end{minipage}
}


\subfigure[Vehicle Point Cloud]{
    \begin{minipage}[t]{0.5\linewidth}
        \centering
        \includegraphics[width=.8\linewidth]{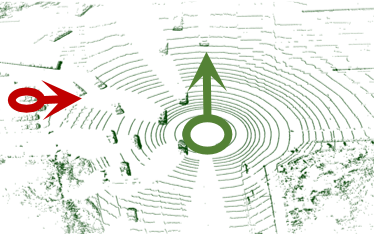}  
        \label{fig:sub3}
    \end{minipage}
}
\hspace{-12mm}
\subfigure[Infrastructure Point Cloud]{
    \begin{minipage}[t]{0.5\linewidth}
        \centering
        \includegraphics[width=.8\linewidth]{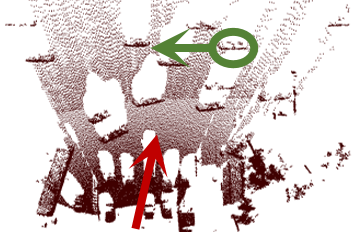}  
        \label{fig:sub4}
    \end{minipage}
}


\subfigure[Cooperative Perspective]{
    \begin{minipage}[t]{0.55\linewidth}
        \centering
        \includegraphics[width=.8\linewidth]{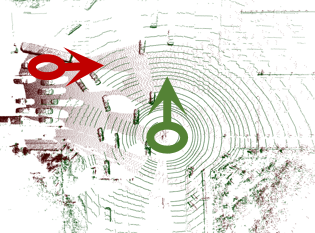}  
        \label{fig:sub5}
    \end{minipage}
}
\hspace{-12mm}
\subfigure{
    \begin{minipage}[t]{0.5\linewidth}
        \centering
        \includegraphics[width=.8\linewidth]{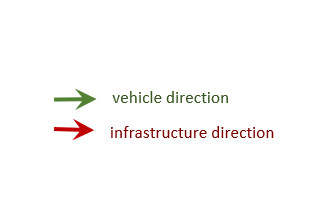}  
        
        \label{fig:sub6}
    \end{minipage}
}

	\caption{Cooperative V2I LiDAR Calibration Schematic. }
	\label{real_test_result}
\end{figure}

To tackle these challenges, the research community has recently seen the emergence of novel approaches. These methods have provided valuable insights into addressing the issue of time synchronization in collaborative vehicle and infrastructure LiDAR systems \cite{SCOPEspatio-temporalICCV2023, SyncNetspatio-temporalECCV2022}, contributing to the enhancement of real-time data transmission and the accuracy of data fusion \cite{FFNetNeurIPS2024}. Despite advancements in spatial alignment, current calibration methods for vehicle-infrastructure cooperative systems typically require initial calibration values as a prerequisite \cite{VI-eyeACM2021, VIPSACM2022}, which can be unstable, especially within city canyons. Moreover, existing studies \cite{lu2023robust} have primarily focused on integrating collaborative calibration methods into cooperative perception strategies, which, to some extent, complicates the model's interpretability. This approach makes it challenging to quantify whether spatial alignment issues within the calibration process are adequately addressed, which is crucial for further enhancing the overall perception results.

A more focused study on LiDAR calibration in cooperative V2X systems should concentrate on the fundamental issues of point cloud registration. Registration can be classified into same-source and cross-source types based on the similarity of data collection devices, or into local and global types depending on initial calibration values.

In essence, the calibration challenge for disparate vehicle and infrastructure LiDAR systems broadly constitutes a cross-source point cloud registration issue \cite{HPCR-VIIV2023}. Furthermore, to maximize the stability of calibration methods within urban canyons, it is advantageous for global point cloud registration to proceed without the necessity for pre-calibration initial values.

Significant progress has been made in same-source point cloud registration \cite{VGICPICRA2021, GMMCVPR2020}, but cross-source registration still faces challenges such as noise, outliers, and variations in density and scale \cite{cross-source-pcarxiv2022}. These challenges necessitate complex computational solutions. Yet, vehicle-infrastructure systems benefit from consistent types of traffic participants in their scenarios, providing a stable foundation for data analysis.

Advancements in global point cloud registration, like those proposed by \cite{qin2023geotransformer, zhou2016fast, yang2020teaser}, offer initial-calibration-free solutions but demand high computational resources, making real-time processing difficult. Current research is thus directed towards developing efficient algorithms, like graph-based optimization techniques \cite{lim2022single}, to reduce computational costs.

To sum up, existing approaches must balance the provision of initial extrinsic values with computational efficiency. This trade-off poses significant challenges for the practical deployment of calibration in V2I cooperative perception systems, necessitating innovative solutions that can bridge the gap between theoretical accuracy and real-world applicability.

This paper presents a novel calibration method for V2I cooperative LiDAR systems. The method establishes an affinity matrix for vehicular and roadside nodes based on the innovative Overall Intersection over Union (oIoU) metric, enabling the identification of shared perception targets between vehicle and infrastructure endpoints. Subsequently, it computes and refines the extrinsic parameters associated with matched boxes. 

The method offers several advantages. Firstly, the proposed method does not require initial extrinsic parameters and meets real-time operation requirements. Secondly, it leverages common target information inherent to traffic scenes, enhancing its generalizability. Furthermore, compared to \cite{HPCR-VIIV2023}, it exclusively utilizes perception information from target detection, resulting in lower computational complexity and reduced data transmission costs, thereby offering greater potential for practical application. Lastly, V2I-Calib is designed with flexibility, having its components well decoupled, which facilitates easy adaptation to meet specific real-world requirements.

The innovations of this paper include:

\begin{itemize}
    \item [1)]      
    We propose a calibration method for cooperative vehicle and infrastructure LiDAR systems that thoroughly exploits the spatial associations between perception objects, characterized by its independence from initial extrinsic parameter values and real-time capabilities.

    \item [2)]
    We introduce the Overall Intersection over Union metric (oIoU), which monitors the real-time calibration performance of extrinsic parameters. Central to this, we propose a method for constructing an affinity matrix for vehicle-road nodes, encoding the correlation between vehicular and roadside targets.
    
    \item [3)]
    The effectiveness of our method is validated through comparative and ablation experiments on the DAIR-V2X dataset.
    
\end{itemize}

\begin{figure*}[htbp] 
\centering 
\includegraphics[width=0.9\textwidth]{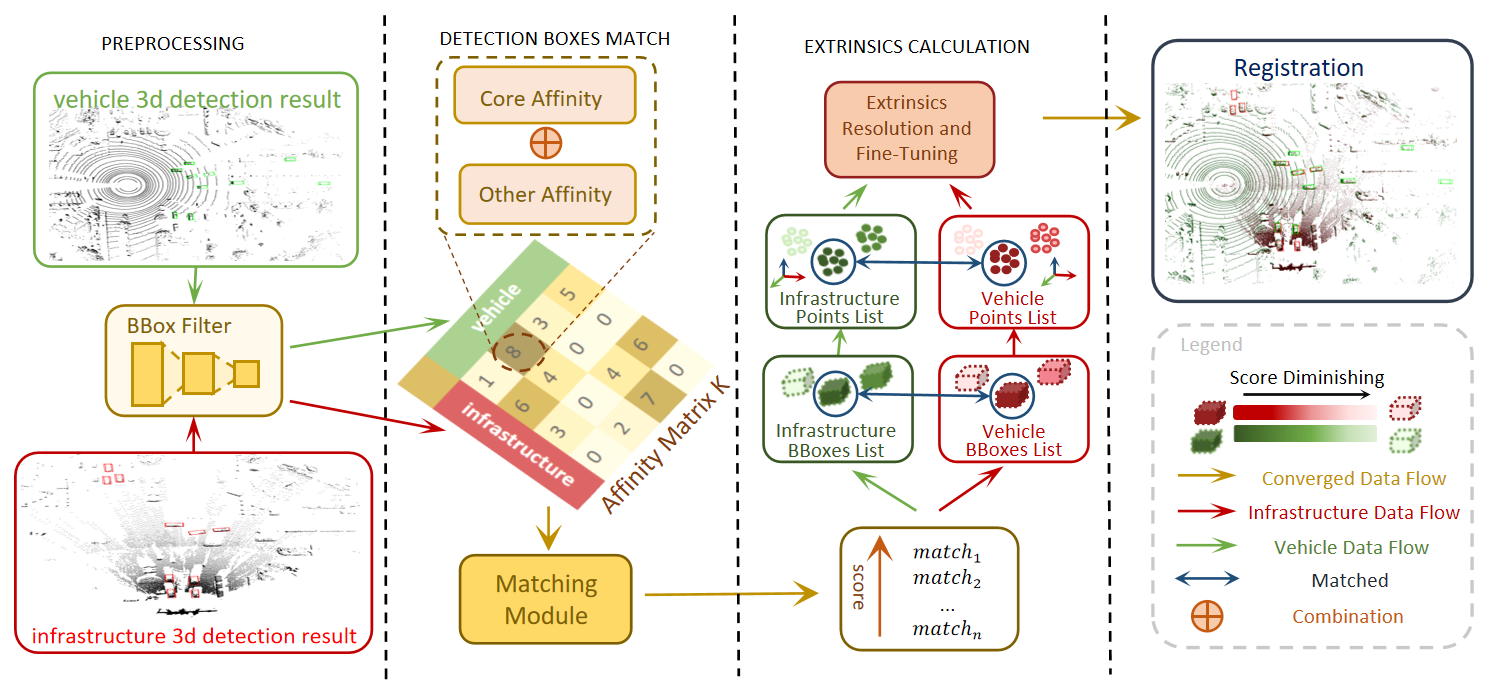} 
\caption{V2I-Calib Workflow. The process initiates with perception objects from vehicle and infrastructure, forming an affinity matrix using Core and Other affinity. Matching algorithms then identify common objects across both ends, facilitating the calculation and optimization of extrinsic parameters.} 
\label{fig:V2I-Calib Workflow} 
\end{figure*}

\section{PROBLEM FORMULATION}

Traditional point cloud registration methods typically seek a rotation matrix $\hat{\mathbf{R}}$ and translation vector $\hat{\mathbf{t}}$ that transform one point cloud $\mathbf{P}_1$ onto another $\mathbf{P}_2$ by minimizing their discrepancy. Mathematically, this is articulated as finding the minimum of an objective function $F(\mathbf{P}_1, \mathbf{P}_2)$, where $F$ is usually a measure of distance between the point clouds:

\begin{equation}
\label{eq:pc_cal}
(\hat{\mathbf{R}}, \hat{\mathbf{t}}) = \arg\min_{\mathbf{R}, \mathbf{t}} F(\mathbf{P}_1^{(c)}, \mathbf{R} \mathbf{P}_2^{(c)} + \mathbf{t})
\end{equation}

However, in the context of vehicle-infrastructure (V2I) cooperation, the point clouds from the vehicle and infrastructure inherently exhibit significant discrepancies, to which the traditional methods may not be directly applicable. In response, this paper introduces an object-aware method. The main concept is to exploit the spatial information of perception objects shared between the vehicle and infrastructure.

Specifically, let $\mathcal{B}^{(\text{veh})}$ denote the set of perception objects from the vehicle, and $\mathcal{B}^{(\text{inf})}$ those from the infrastructure. Our objective is to identify the common perception object set $\mathcal{B}^{(c)}$, comprising the common targets detected by both systems. Mathematically, the common detection box set is expressed as:

\begin{equation}
\mathcal{B}^{(c)} = \mathcal{B}^{(veh)} \cap \mathcal{B}^{(inf)} 
\end{equation}

After extracting common perception objects from the original point cloud, we transform these objects into feature point clouds, denoted as $\mathbf{P}^{(\text{c})}$. This transformation involves mapping the vertices of the perception objects into point sets, a process facilitated by the function $\text{Vertices}(\cdot)$. Specifically, the vertices of each detection box $\mathcal{B}_{i}$ are extracted and merged to form the complete feature point cloud:

\begin{equation}
\label{eq:box2point}
\mathbf{P}^{(c)} = \bigcup_{\mathcal{B}_{i} \in \mathcal{B}^{(c)}} \text{Vertices}(\mathcal{B}_{i})
\end{equation}

This transformation yields a well-characterized feature point cloud, which can then be processed within the framework of classical point cloud registration algorithms, as formulated in Equation (\ref{eq:pc_cal}). Upon obtaining the rotation matrix $\hat{\mathbf{R}}$ and translation vector $\hat{\mathbf{t}}$, these are integrated to produce the optimal extrinsic parameter result $\hat{\mathbf{T}}$.

\begin{equation}
\hat{\mathbf{T}} = \begin{bmatrix}
\hat{\mathbf{R}} & \hat{\mathbf{t}} \\
\mathbf{0}^{T} & 1
\end{bmatrix}
\end{equation}

By employing the point cloud registration method based on common perception objects, we can effectively tackle the challenges of cross-source point cloud registration in V2I cooperative scenarios. The crux of the issue now becomes how to identify the common detection box set $\mathcal{B}^{(c)}$ and then fine-tune the extrinsics, which will be elaborated in subsequent chapters.

\section{METHODOLOGY}
\label{section:methodology}



In this study, we introduce a novel vehicle-infrastructure LiDAR calibration method (V2I-Calib), the overall workflow of which is illustrated in Figure \ref{fig:V2I-Calib Workflow}. 


\subsection{Affinity Formulation} 
\label{affinity formulation}


In this section, we explore the development of affinity measures critical to the vehicle-infrastructure cooperative LiDAR calibration framework. By creating a detailed affinity function, we effectively assess the alignment between vehicle-side and infrastructure-side perception objects, essential for the next matching module. The challenges include differing coordinate systems for perception objects and the randomness in cross-source point cloud data, complicating the identification of shared versus unique detection objects. We address these challenges using a hypothesized matching pairs strategy and the Overall Intersection over Union metric(oIoU). We'll discuss the formulation of affinity measures in two parts: core affinity and additional affinities.

\subsubsection{Core Affinity}

This section outlines the formulation of Core Affinity by evaluating scene matching scores using the Overall Intersection over Union (oIoU) metric, which measures the extent of scene congruence based on detection box pairs from both vehicle and infrastructure sources:

\begin{equation}
   \emph{oIoU} = \frac{1}{\max(m, n)} \sum_{i=0}^{\mathbf{m}-1} \sum_{j=0}^{\mathbf{n}-1} \emph{IoU}_{3D}(\mathcal{B}^{(inf)}_{i}, \mathcal{B}^{(veh)}_{j})
\end{equation}

\begin{equation}
    IoU_{3D}(\mathcal{A}, \mathcal{B}) = \frac{\text{Vol}(\mathcal{A} \cap \mathcal{B})}{\text{Vol}(\mathcal{A} \cup \mathcal{B})}
\end{equation}

In this model, $m$ and $n$ are the counts of 3D detection boxes by the infrastructure and vehicle, respectively. $\emph{IoU}_{3D}(\cdot)$ calculates volumetric overlaps in 3D space, where \( \text{Vol}(\cdot) \) denotes the volumetric overlap between two 3d detection boxes. oIoU advances beyond traditional IoU by incorporating positional relationships among boxes, thus providing a comprehensive evaluation of spatial alignment.

\begin{figure}[tbp] 
\centering 
\includegraphics[width=2.5in]{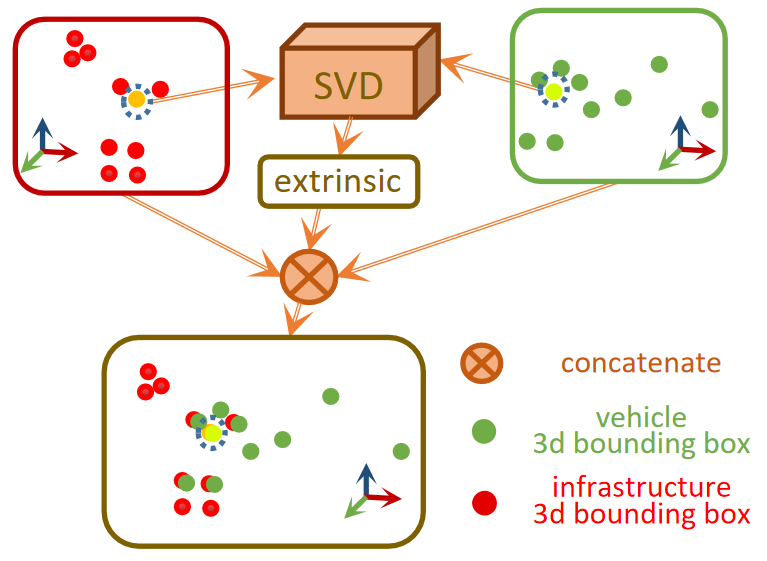} 
\caption{SVD-based extrinsic parameters are computed using selected 3D detection boxes from vehicle (green) and infrastructure (red) as assumed matching pairs \cite{SVDPAMI1987}. The scene is transformed and merged to assess scene affinity via the oIoU metric.} 
\label{Core Affinity} 
\end{figure}

Building upon this foundation, the process of constructing Core Affinity, as illustrated in Figure \ref{Core Affinity}, employs the concept of hypothetical matching pairs to define the feasible range of extrinsic parameters search between the vehicle and the infrastructure, as per the method outlined in \cite{SVDPAMI1987}. The oIoU metric is then applied to assess the quality of spatial alignment for the corresponding extrinsic parameters of these tentative vehicle-infrastructure matching pairs.

\subsubsection{Additional Affinity Calculation}

Beyond the core affinity, this section delves into additional affinity metrics aimed at bolstering the precision and durability of the matching process. One such metric is category affinity, which evaluates the congruence of categories between detection boxes on the vehicle and infrastructure sides in the assumed matching pairs. This consideration is pivotal in diminishing the computational demands associated with core affinity, by ensuring that only detection boxes of matching categories are compared, thereby streamlining the matching process and enhancing its efficiency. 

Furthermore, affinities such as size affinity, tracking ID affinity, angle affinity, and length affinity from VIPS\cite{VIPSACM2022}, as well as appearance affinity from DeepSORT\cite{DeepSORTICIP2017}, can serve as supplements specific to scenarios to enhance the calibration stability of vehicle-infrastructure LiDAR systems in environments where shared objects are sparse.

In general scenarios, a combination of Core Affinity and Category Affinity proves to be effective, as demonstrated in section \ref{section:experiment}.

\subsection{Affinity Matrix Formulation and Matching}


This module constructs an affinity matrix from vehicle and infrastructure detection boxes, treating them as nodes within weighted graphs \emph{$G_{\text{veh}}$} and \emph{$G_{\text{inf}}$}. This setup transforms the matching problem into a graph matching problem \cite{GM-surveyICISP2020}, often modeled as a quadratic assignment problem (QAP) \cite{FGMTPAMI2015}, involving vertex affinity matrix \( \mathbf{K}_p \) and the edge affinity matrix \( \mathbf{K}_q \).

Due to the NP-Hard nature of QAP and its high computational demand, we simplify the problem to a linear assignment format by integrating \( \mathbf{K}_p \) and \( \mathbf{K}_q \) into a single affinity matrix \( \mathbf{A} \):

\begin{equation} 
\mathbf{A}(i, j) = \alpha \cdot \mathbf{K}_p(i, j) + \beta \cdot \sum_{u \in E(i), v \in E(j)} \mathbf{K}_q(u, v)
\end{equation}

Here, \( E(i) \) represents the set of all edges connected to node \( i \). The parameters \( \alpha \) and \( \beta \) are the adjustment factors for the vertex affinity \( \mathbf{K}_p(i, j) \) and edge affinity \( \mathbf{K}_q(i, j) \), respectively, and are empirically set to 0.5. The summation aggregates the edge affinities to complement the vertex affinity in the matrix \( \mathbf{A} \).

With affinity matrix \( \mathbf{A} \), the Hungarian method \cite{HungarianNRLQ1995} efficiently finds optimal matches in polynomial time while reducing computational complexity, suitable for the dynamics of vehicle-infrastructure collaboration.

\subsection{Extrinsics Resolution and Optimization}
\label{extrinsics optimization}



The objective of this section is to compute extrinsic parameters from matched detection box pairs and to ensure their accuracy, addressing two primary challenges: the unknown correspondence of point pairs within the boxes, influenced by the rotation matrix of the extrinsics, and the ill-posed nature of calibration that requires a good initial estimate to balance accuracy with computational cost.

After obtaining \( \mathcal{B}^{(c)} \), we derive the feature point clouds \( \mathbf{P}^{\text{(inf)}} \) and \( \mathbf{P}^{\text{(veh)}} \) using Equation (\ref{eq:box2point}), transforming the initial cross-source point cloud calibration challenge into a standardized point set registration problem.




A naive extrinsic parameter calculation method provides an initial value \cite{SVDPAMI1987}, which is optimized to derive the optimal parameters \cite{ICPTPAMI1992, NDTIROS2003}.


\section{EXPERIMENT} 
\label{section:experiment}

\subsection{Evaluation Metrics}

The core metrics for experimental evaluation include Relative Rotation Error (RRE), Relative Translation Error (RTE), and Success Rate.

\textbf{Relative Rotation Error (RRE):} Measures the accuracy of the rotation component in the calibration result, i.e., the angular difference between the estimated rotation matrix \( \mathbf{R}_e \) and the ground truth \( \mathbf{R}_t \) .


\begin{equation}
    \emph{RRE} = \arccos\left(\frac{\text{tr}(\mathbf{\Delta R}) - 1}{2}\right) 
\end{equation}
\begin{equation}
    \mathbf{\Delta R} = \mathbf{R}_t^{-1}\mathbf{R}_e
\end{equation}

\textbf{Relative Translation Error (RTE):} Evaluates the accuracy of the translation vector in the calibration result, i.e., the distance difference between the estimated translation vector \( \mathbf{t}_e \) and the ground truth translation vector \( \mathbf{t}_t \).

\begin{equation}
    \emph{RTE} = || \mathbf{t}_t - \mathbf{t}_e ||_2
\end{equation}

\textbf{Success Rate:} Defined as the proportion of successfully completed calibration tasks within a preset error threshold, reflecting the method's robustness and reliability. Following the criteria in \cite{VIPSACM2022}, we consider $\emph{RTE} < 2m$ as the determinant criterion.

These metrics can provide information on the adaptability and stability of the method across different scenarios.

\subsection{Datasets}

\begin{figure*}[htbp]
	\centering
	\setcounter{subfigure}{0} 
	
	\subfigure[Ideal Scenario]{
		\begin{minipage}[t]{0.23\linewidth}
			\centering
			\includegraphics[width=\linewidth]{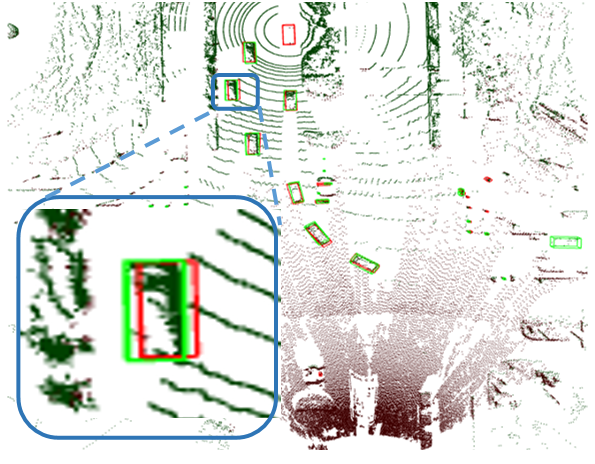}
			\label{fig:testresult:sub1}
		\end{minipage}
	}
	\subfigure[Challenging Scenario]{
		\begin{minipage}[t]{0.23\linewidth}
			\centering
			\includegraphics[width=\linewidth]{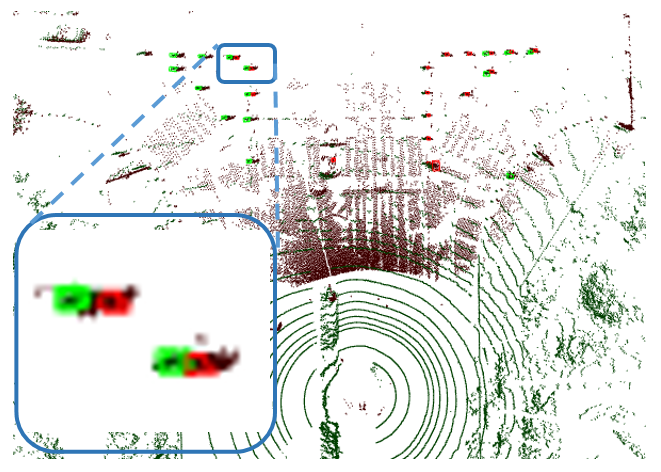}
			\label{fig:testresult:sub2}
		\end{minipage}
	}
	\subfigure[Complex Scenario]{
		\begin{minipage}[t]{0.23\linewidth}
			\centering
			\includegraphics[width=\linewidth]{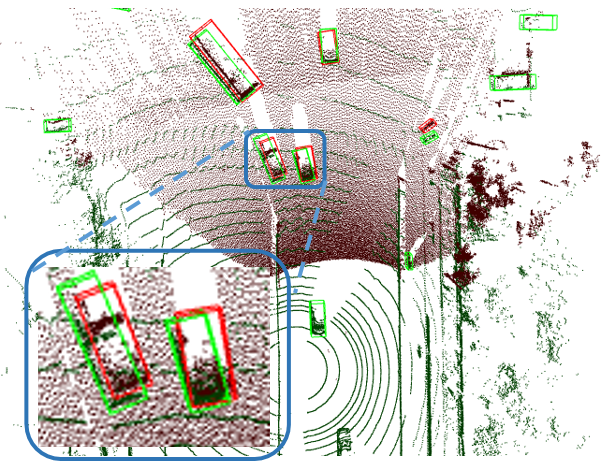}
			\label{fig:testresult:sub3}
		\end{minipage}
	}
	\subfigure[Complex Scenario]{
		\begin{minipage}[t]{0.23\linewidth}
			\centering
			\includegraphics[width=\linewidth]{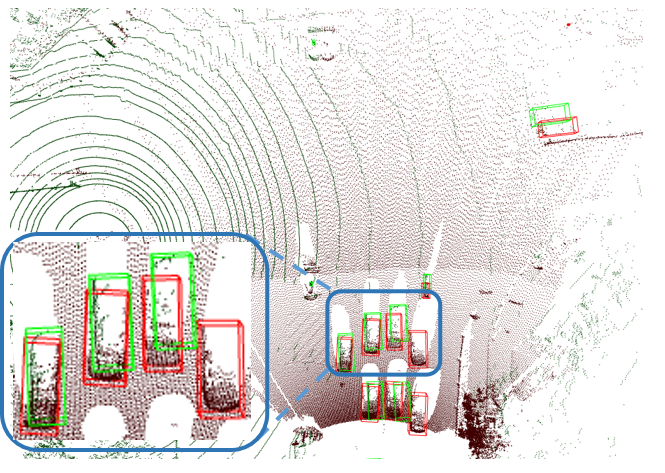}
			\label{fig:testresult:sub4}
		\end{minipage}
	}
	
	\caption{Comparative Calibration Results Across Diverse Scenarios: (a) Accurate alignment of vehicle and infrastructure detection boxes. (b) Calibration errors due to small-sized detection boxes. (c, d) Good and Suboptimal calibration results combining multiple detection boxes, respectively. } 
	\label{fig:testresult} 
\end{figure*}

In this study, we conducted experimental validation using the DAIR-V2X dataset\cite{dairv2xCVPR2023}, which encompasses a wealth of data collected from Vehicle-Infrastructure Cooperative Autonomous Driving (VICAD) scenarios, including vehicular and infrastructural LiDAR data along with their 3D box annotations. The specifications of the vehicular and roadside LiDAR systems are presented in Table \ref{tab:lidar_specs}. The outcomes of the experimental scenarios are illustrated in Figure \ref{fig:testresult}.

\begin{table}[h]
\centering
\caption{Specifications of LiDAR Equipment}
\label{tab:lidar_specs}
\renewcommand{\arraystretch}{1.5} 
\begin{tabular}{
  >{\centering\arraybackslash}m{0.35\columnwidth} 
  >{\centering\arraybackslash}m{0.25\columnwidth}
  >{\centering\arraybackslash}m{0.25\columnwidth}
}
\hline
\textbf{Parameter} & \textbf{Roadside LiDAR} & \textbf{Vehicle LiDAR} \\
\hline
LiDAR Points & 300 lines & 40 lines \\
Horizontal Field of View & 100° & 360° \\
Max Detection Range & 280 meters & 200 meters \\
\hline
\end{tabular}
\end{table}

\subsection{Experimental Settings}

This section describes a systematic performance evaluation of the proposed method, which is conducted through comparative experiments and ablation studies.All experiments were performed on an experimental platform equipped with an Intel i7-9750H CPU. 

\subsubsection{Contrast Experiment}

The comparative experiments assessed the performance of the proposed method against other well-performing, global registration methods \cite{zhou2016fast, yang2020teaser, lim2022single}.  

\begin{table}[htbp]
\centering
\caption{ Comparative Results of Different Methods.}
\footnotesize 
\setlength{\tabcolsep}{4pt} 
\renewcommand{\arraystretch}{1.1} 
\begin{tabular}{@{}>{\centering\arraybackslash}m{1.3cm} >{\centering\arraybackslash}m{1.8cm} >{\centering\arraybackslash}m{1cm} >{\centering\arraybackslash}m{1cm} >{\centering\arraybackslash}m{1cm} >{\centering\arraybackslash}m{1cm}@{}}
\toprule
\textbf{Dataset} \newline \textbf{Difficulty} & \textbf{Methods} & \textbf{RRE(°)} & \textbf{RTE(m)} & \textbf{Success} \newline \textbf{Rate(\%)} & \textbf{Time (s)} \\ 
\midrule
\multirow{5}{*}{easy group} 
& FGR \cite{zhou2016fast} & 0.98 & 0.93 & 59.8 & 22.23 \\
& Teaser++ \cite{yang2020teaser} & 0.99 & 0.93 & 60.4 & 22.74 \\
& Quatro \cite{lim2022single} & 1.05 & 1.04 & 49.5 & 23.37 \\
& \textbf{V2I-Calib} & \textbf{0.68} & \textbf{0.56} & \textbf{96.8} & \textbf{0.21} \\
\midrule
\multirow{5}{*}{hard group} 
& FGR \cite{zhou2016fast} & \textbf{1.15} & 1.11 & 24.8 & 15.17 \\
& Teaser++ \cite{yang2020teaser} & 1.17 & \textbf{1.10} & 25.7 & 14.79 \\
& Quatro \cite{lim2022single} & 1.23 & 1.21 & 19.2 & 16.67 \\
& \textbf{V2I-Calib} & 1.92 & 1.67 & \textbf{71.8} & \textbf{0.15} \\
\bottomrule
\end{tabular}
\label{tab:constrast experiment}
\end{table}

As indicated in Table \ref{tab:constrast experiment}, it is evident that the best current global registration methods struggle with the time complexity required for real-time processing when facing large-scale point clouds such as those in the DAIR-V2X dataset. Although these methods achieve near 50\% success rate and obtain reasonably accurate initial calibration values on the dataset, they lack practical applicability. 
In contrast, the proposed V2I-Calib method significantly outperforms existing global point cloud registration algorithms in terms of runtime and achieves considerably higher accuracy in simple scenarios with lower complexity. In hard groups with more occlusions, the precision of our method is impacted due to the lower accuracy of detection results. Notably, the running times for the methods in the more challenging group are generally lower than those in the simpler group, which is attributable to the fewer shared targets in hard scenes, reducing the computational load.

	
	

\subsubsection{Ablation experiment}

In this section, we aim to validate the effectiveness of the core affinity formulation module presented in Section \ref{affinity formulation} and the extrinsic parameter optimization module discussed in Section \ref{extrinsics optimization} through a series of ablation studies.

\begin{table}[htbp]
\centering
\caption{Methods Composition Of Different Strategies.}
\begin{tabular}{lcc}
\toprule
\textbf{Method} & \textbf{Affinity Construction} & \textbf{Extrinsic Optimization} \\
\midrule
V2I-Calib-v3 & Length and Angle \cite{VIPSACM2022} & Yes \\
V2I-Calib-v2 & Core and Category & No \\
V2I-Calib-v1 & Core and Category & Yes \\
\bottomrule
\end{tabular}
\label{tab:method_comparison}
\end{table}

\begin{table}[htbp]
\centering
\caption{Ablation Study On Methods With Different Strategies}
\footnotesize 
\setlength{\tabcolsep}{4pt} 
\renewcommand{\arraystretch}{1.1} 
\begin{tabular}{@{}>{\centering\arraybackslash}m{1.3cm} >{\centering\arraybackslash}m{1.8cm} >{\centering\arraybackslash}m{1cm} >{\centering\arraybackslash}m{1cm} >{\centering\arraybackslash}m{1cm} >{\centering\arraybackslash}m{1cm}@{}}
\toprule
\textbf{Dataset} \newline \textbf{Difficulty} & \textbf{Methods} & \textbf{RRE(°)} & \textbf{RTE(m)} & \textbf{Success} \newline \textbf{Rate(\%)} & \textbf{Time (s)} \\ 
\midrule
\multirow{3}{*}{easy group} 
& V2I-Calib-v3 & 0.95 & 0.79 & 80.3 & 0.32 \\
& V2I-Calib-v2 & 0.98 & 0.62 & 50.0 & \textbf{0.18} \\
& \textbf{V2I-Calib-v1} & \textbf{0.68} & \textbf{0.56} & \textbf{96.8} & 0.21 \\
\midrule
\multirow{3}{*}{hard group} 
& V2I-Calib-v3 & 2.69 & 2.37 & 57.2 & 0.28 \\
& V2I-Calib-v2 & 1.93 & 1.82 & 53.4 & \textbf{0.14} \\
& \textbf{V2I-Calib-v1} & \textbf{1.92} & \textbf{1.67} & \textbf{71.8} & 0.15 \\
\bottomrule
\end{tabular}
\label{tab:calibration_comparison}
\end{table}

\begin{figure}[htbp] 
\centering 
\includegraphics[width=0.4\textwidth]{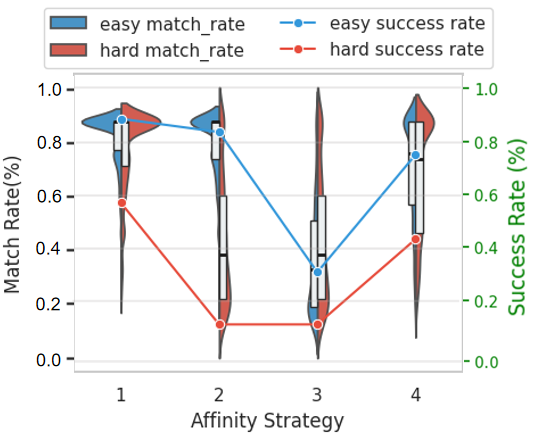} 
\caption{Ablation Comparison of Affinity Strategies. Strategy 1 combines core and category affinities; Strategy 2 uses core affinity alone; Strategy 3 combines angle and category affinities; Strategy 4 integrates length with category affinities.}
\label{fig:violin} 
\end{figure}

As indicated in Tables \ref{tab:method_comparison} and \ref{tab:calibration_comparison}, the core affinity formulation module introduced in this study significantly enhances robustness, evidenced by a 20\% increase in success rates when comparing V2I-Calib-v1 with V2I-Calib-v3. Furthermore, the inclusion of the extrinsic parameter optimization module, as evaluated through a comparison of V2I-Calib-v1 and V2I-Calib-v2, substantially improves calibration precision and success rates.

The superiority of the proposed affinity metrics was validated against category, angle, and length affinities, detailed in \cite{VIPSACM2022}. The results, depicted in Figure \ref{fig:violin}, show that Strategy 1 (core affinity combined with category affinity) not only achieves higher success rates and accuracy but also reduces computational time significantly compared to Strategy 2 (core affinity alone), especially in challenging scenarios.

Regarding the Strategy 3 (angle affinity with category affinity) and Strategy 4 (length affinity with category affinity) extended from \cite{VIPSACM2022}, these strategies were originally designed for scenarios assisted by positional information. However, experimental results show that they are somewhat applicable in scenarios without initial extrinsic parameter values, although their performance significantly decreases in challenging settings. Upon further analysis of the scenarios where these strategies failed, we observed that most of these scenarios involved extrinsic parameters with relatively large absolute values, aligning with the limitations inherent in their initial design rationale.


\section{CONCLUSIONS}
\label{section:conclusions}

This study introduces V2I-Calib, a calibration algorithm tailored for vehicle-infrastructure cooperative LiDAR systems. Unlike existing techniques, it does not require initial external parameters, enhancing stability in practical applications. Additionally, V2I-Calib satisfies real-time performance criteria, setting it apart from currently effective global point cloud registration methods.

Future research will focus on integrating temporal synchronization tasks to enhance practicality, alongside incorporating 2D bounding box data to improve multi-sensor fusion calibration. As the technology evolves, the V2I-Calib framework is expected to adapt to broader applications, enhancing the safety and reliability of autonomous driving systems.







\begin{thebibliography}{99}

\bibitem{autonomousdrivingIROS2023}
C.~Brewitt, M.~Tamborski, C.~Wang, and S.~V. Albrecht, ``Verifiable goal recognition for autonomous driving with occlusions,'' in \emph{2023 IEEE/RSJ International Conference on Intelligent Robots and Systems (IROS)}.\hskip 1em plus 0.5em minus 0.4em\relax IEEE, 2023, pp. 11\,210--11\,217.

\bibitem{badweathersensors2024}
J.~Wang, Z.~Wu, Y.~Liang, J.~Tang, and H.~Chen, ``Perception methods for adverse weather based on vehicle infrastructure cooperation system: A review,'' \emph{Sensors}, vol.~24, no.~2, p. 374, 2024.

\bibitem{cooperativeIROS2023}
B.~Cao, C.-N. Ritter, K.~Alomari, and D.~Goehring, ``Cooperative lidar localization and mapping for v2x connected autonomous vehicles,'' in \emph{2023 IEEE/RSJ International Conference on Intelligent Robots and Systems (IROS)}.\hskip 1em plus 0.5em minus 0.4em\relax IEEE, 2023, pp. 11\,019--11\,026.

\bibitem{SCOPEspatio-temporalICCV2023}
K.~Yang, D.~Yang, J.~Zhang, M.~Li, Y.~Liu, J.~Liu, H.~Wang, P.~Sun, and L.~Song, ``Spatio-temporal domain awareness for multi-agent collaborative perception,'' in \emph{Proceedings of the IEEE/CVF International Conference on Computer Vision}, 2023, pp. 23\,383--23\,392.

\bibitem{SyncNetspatio-temporalECCV2022}
Z.~Lei, S.~Ren, Y.~Hu, W.~Zhang, and S.~Chen, ``Latency-aware collaborative perception,'' in \emph{European Conference on Computer Vision}.\hskip 1em plus 0.5em minus 0.4em\relax Springer, 2022, pp. 316--332.

\bibitem{FFNetNeurIPS2024}
H.~Yu, Y.~Tang, E.~Xie, J.~Mao, P.~Luo, and Z.~Nie, ``Flow-based feature fusion for vehicle-infrastructure cooperative 3d object detection,'' \emph{Advances in Neural Information Processing Systems}, vol.~36, 2024.

\bibitem{VI-eyeACM2021}
Y.~He, L.~Ma, Z.~Jiang, Y.~Tang, and G.~Xing, ``Vi-eye: semantic-based 3d point cloud registration for infrastructure-assisted autonomous driving,'' in \emph{Proceedings of the 27th Annual International Conference on Mobile Computing and Networking}, 2021, pp. 573--586.

\bibitem{VIPSACM2022}
S.~Shi, J.~Cui, Z.~Jiang, Z.~Yan, G.~Xing, J.~Niu, and Z.~Ouyang, ``Vips: Real-time perception fusion for infrastructure-assisted autonomous driving,'' in \emph{Proceedings of the 28th Annual International Conference on Mobile Computing And Networking}, 2022, pp. 133--146.

\bibitem{lu2023robust}
Y.~Lu, Q.~Li, B.~Liu, M.~Dianati, C.~Feng, S.~Chen, and Y.~Wang, ``Robust collaborative 3d object detection in presence of pose errors,'' in \emph{2023 IEEE International Conference on Robotics and Automation (ICRA)}.\hskip 1em plus 0.5em minus 0.4em\relax IEEE, 2023, pp. 4812--4818.

\bibitem{HPCR-VIIV2023}
Y.~Zhao, X.~Zhang, S.~Zhang, S.~Qiu, H.~Yin, and X.~Zhang, ``Hpcr-vi: Heterogeneous point cloud registration for vehicle-infrastructure collaboration,'' in \emph{2023 IEEE Intelligent Vehicles Symposium (IV)}.\hskip 1em plus 0.5em minus 0.4em\relax IEEE, 2023, pp. 1--6.

\bibitem{VGICPICRA2021}
K.~Koide, M.~Yokozuka, S.~Oishi, and A.~Banno, ``Voxelized gicp for fast and accurate 3d point cloud registration,'' in \emph{2021 IEEE International Conference on Robotics and Automation (ICRA)}.\hskip 1em plus 0.5em minus 0.4em\relax IEEE, 2021, pp. 11\,054--11\,059.

\bibitem{GMMCVPR2020}
A.~Hertz, R.~Hanocka, R.~Giryes, and D.~Cohen-Or, ``Pointgmm: A neural gmm network for point clouds,'' in \emph{Proceedings of the IEEE/CVF Conference on Computer Vision and Pattern Recognition}, 2020, pp. 12\,054--12\,063.

\bibitem{cross-source-pcarxiv2022}
X.~Huang, G.~Mei, and J.~Zhang, ``Cross-source point cloud registration: Challenges, progress and prospects,'' \emph{Neurocomputing}, p. 126383, 2023.

\bibitem{qin2023geotransformer}
Z.~Qin, H.~Yu, C.~Wang, Y.~Guo, Y.~Peng, S.~Ilic, D.~Hu, and K.~Xu, ``Geotransformer: Fast and robust point cloud registration with geometric transformer,'' \emph{IEEE Transactions on Pattern Analysis and Machine Intelligence}, vol.~45, no.~8, pp. 9806--9821, 2023.

\bibitem{zhou2016fast}
Q.-Y. Zhou, J.~Park, and V.~Koltun, ``Fast global registration,'' in \emph{Computer Vision--ECCV 2016: 14th European Conference, Amsterdam, The Netherlands, October 11-14, 2016, Proceedings, Part II 14}.\hskip 1em plus 0.5em minus 0.4em\relax Springer, 2016, pp. 766--782.

\bibitem{yang2020teaser}
H.~Yang, J.~Shi, and L.~Carlone, ``Teaser: Fast and certifiable point cloud registration,'' \emph{IEEE Transactions on Robotics}, vol.~37, no.~2, pp. 314--333, 2020.

\bibitem{lim2022single}
H.~Lim, S.~Yeon, S.~Ryu, Y.~Lee, Y.~Kim, J.~Yun, E.~Jung, D.~Lee, and H.~Myung, ``A single correspondence is enough: Robust global registration to avoid degeneracy in urban environments,'' in \emph{2022 international conference on robotics and automation (ICRA)}.\hskip 1em plus 0.5em minus 0.4em\relax IEEE, 2022, pp. 8010--8017.

\bibitem{SVDPAMI1987}
K.~S. Arun, T.~S. Huang, and S.~D. Blostein, ``Least-squares fitting of two 3-d point sets,'' \emph{IEEE Transactions on pattern analysis and machine intelligence}, no.~5, pp. 698--700, 1987.

\bibitem{DeepSORTICIP2017}
N.~Wojke, A.~Bewley, and D.~Paulus, ``Simple online and realtime tracking with a deep association metric,'' in \emph{2017 IEEE international conference on image processing (ICIP)}.\hskip 1em plus 0.5em minus 0.4em\relax IEEE, 2017, pp. 3645--3649.

\bibitem{GM-surveyICISP2020}
H.~Sun, W.~Zhou, and M.~Fei, ``A survey on graph matching in computer vision,'' in \emph{2020 13th International Congress on Image and Signal Processing, BioMedical Engineering and Informatics (CISP-BMEI)}.\hskip 1em plus 0.5em minus 0.4em\relax IEEE, 2020, pp. 225--230.

\bibitem{FGMTPAMI2015}
F.~Zhou and F.~De~la Torre, ``Factorized graph matching,'' \emph{IEEE transactions on pattern analysis and machine intelligence}, vol.~38, no.~9, pp. 1774--1789, 2015.

\bibitem{HungarianNRLQ1995}
H.~W. Kuhn, ``The hungarian method for the assignment problem,'' \emph{Naval research logistics quarterly}, vol.~2, no. 1-2, pp. 83--97, 1955.

\bibitem{ICPTPAMI1992}
P.~J. Besl and N.~D. McKay, ``Method for registration of 3-d shapes,'' in \emph{Sensor fusion IV: control paradigms and data structures}, vol. 1611.\hskip 1em plus 0.5em minus 0.4em\relax Spie, 1992, pp. 586--606.

\bibitem{NDTIROS2003}
P.~Biber and W.~Stra{\ss}er, ``The normal distributions transform: A new approach to laser scan matching,'' in \emph{Proceedings 2003 IEEE/RSJ International Conference on Intelligent Robots and Systems (IROS 2003)(Cat. No. 03CH37453)}, vol.~3.\hskip 1em plus 0.5em minus 0.4em\relax IEEE, 2003, pp. 2743--2748.

\bibitem{dairv2xCVPR2023}
H.~Yu, W.~Yang, H.~Ruan, Z.~Yang, Y.~Tang, X.~Gao, X.~Hao, Y.~Shi, Y.~Pan, N.~Sun, \emph{et~al.}, ``V2x-seq: A large-scale sequential dataset for vehicle-infrastructure cooperative perception and forecasting,'' in \emph{Proceedings of the IEEE/CVF Conference on Computer Vision and Pattern Recognition}, 2023, pp. 5486--5495.


\end{thebibliography}

\end{document}